\documentclass[journal]{IEEEtran}

\ifCLASSOPTIONcompsoc
\else

\usepackage{amsmath,epsfig,epstopdf}
\usepackage{amsfonts}
\usepackage{times}
\usepackage{amssymb}
\usepackage{graphicx}
\usepackage{euscript}
\usepackage{algorithm, algorithmic}
\usepackage{color}
\usepackage{multirow,hhline,colortbl}
\usepackage{booktabs}
\usepackage{balance}

\fi
\ifCLASSINFOpdf
\else
\fi

\newcommand{\ru}    {\rule{0mm}{4mm}}

\newcommand{\be}    {\begin{equation}}
\newcommand{\ee}    {\end{equation}}

\newcommand{\sep}{\!\!&\!\!}

\begin{document}
\title{Land Use Classification in Remote Sensing Images by Convolutional Neural Networks}

\author{Marco~Castelluccio, Giovanni~Poggi, Carlo~Sansone, Luisa~Verdoliva
\IEEEcompsocitemizethanks{
\IEEEcompsocthanksitem The Authors
are with the DIETI, Universit\`{a} Federico II di Napoli, Naples, Italy. \protect
E-mail: mar.castelluccio@studenti.unina.it, \{poggi, carlosan, verdoliv\}@unina.it} }

\markboth{}%
{Castelluccio \MakeLowercase{\textit{et al.}}: ....}
\IEEEcompsoctitleabstractindextext{%

\begin{abstract}
We explore the use of convolutional neural networks for the semantic classification of remote sensing scenes.
Two recently proposed architectures, CaffeNet and GoogLeNet, are adopted, with three different learning modalities.
Besides conventional training from scratch,
we resort to pre-trained networks that are only fine-tuned on the target data, so as to avoid overfitting problems and reduce design time.
Experiments on two remote sensing datasets, with markedly different characteristics,
testify on the effectiveness and wide applicability of the proposed solution,
which guarantees a significant performance improvement over all state-of-the-art references.
\end{abstract}

}
\maketitle

\IEEEdisplaynotcompsoctitleabstractindextext
\IEEEpeerreviewmaketitle

\section{Introduction}
Thanks to the rapid progresses in remote sensing technology, and the reduction of acquisition costs,
a large bulk of images of the Earth is readily available nowadays.
They are taken from satellites or airplanes, with various imaging modalities, spatial and spectral resolutions, dynamic ranges.

With no shortage of data,
the focus shifts on the ability to automatically extract valuable information from them.
In recent years,
there have been great advances in remote sensing image processing,
for both low-level tasks, such as denoising or segmentation, and high level ones, such as classification.
A plethora of land cover classification algorithms have been developed, with solid theoretical foundations,
based on spectral and spatial properties of the pixels.
However, the task becomes incrementally more difficult as the level of abstraction increases,
going from pixels, to objects, and then scenes.

\begin{figure}
\centering
\begin{tabular}{cc}
\includegraphics[width=.40\columnwidth]{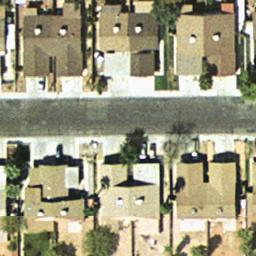}&
\includegraphics[width=.40\columnwidth]{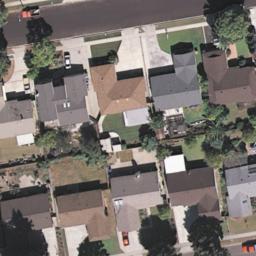}\\
(a)&(b)\\
\end{tabular}
\caption{Too close to call?
Two similar images from the dense residential (a) and medium residential (b) classes of the UC-Merced dataset.
Notice the different scales of observation.}
\label{fig:tooclosetocall}
\end{figure}

Labeling an image according to a set of semantic categories is the goal of scene classification.
This is a very challenging problem,
because land covers characterizing a given class may present a large variability and objects may appear at different scales and orientations.
High intra-class variability then couples with low inter-class distance,
a problem that grows ever more as finer classifications are sought.
The same land covers and even the same objects can be found in images belonging to different classes.
An example is shown in Fig.\ref{fig:tooclosetocall}
where the difference is made only by the density of buildings.

In this context,
low-level features typical of pixel-based or object-based approaches \cite{Pesaresi2011, Rizvi2011, Gaetano2015},
encoding spectral, textural, and geometrical properties, become mostly ineffective.
More complex features and descriptors are necessary to capture the semantics of the scene.
These have been the object of intense research efforts in the last few years, leading to good results in many fields.
Nonetheless, even these sophisticated and successful descriptors are rapidly giving way to deep neural networks.

Artificial neural networks take inspiration from models of the biological brain,
and try to reproduce some of its functions by using simple but massively interconnected processing units, the neurons.
A typical neural network architecture comprises several layers of neurons feeding one another, by which the ``deep'' attribute.
Deep learning has provided impressive results in object recognition \cite{Krizhevsky2012}.
Recently, it has been also applied to remote sensing tasks \cite{Midhun2014} \cite{Chen2015b} \cite{Li2014},
including land use image classification \cite{Penatti2015}, showing always a great potential.

Considering the subtle differences among categories in scene classification,
the superiority of deep learning with respect to ``shallow'' descriptors mentioned before can be easily claimed.
While the latter aim at {\em reproducing the behavior} of the human interpreter, associating labels to images through a black box,
deep neural networks try to {\em replicate cerebral mechanisms},
learning and combining internal descriptions of increasing levels of abstraction.
That this be actually the case may be the object of endless controversies.
Nonetheless, performance figures speak consistently in favor of deep learning for such tasks.

This work adds another piece of evidence in this sense.
We use deep convolutional neural networks (CNNs or ConvNets) to tackle the remote sensing scene classification task.
Two recently proposed promising architectures, CaffeNet \cite{Jia2014} and GoogLeNet \cite{Szegedy2015}, are considered and tested.
To cope with the scarcity of remote sensing training data, we explore various training modalities:
not only the usual training from scratch but also the fine-tuning of pre-trained networks.
Experiments on two publicly available remote sensing datasets, with widely different characteristics,
prove CNNs to provide always an excellent performance.
On the well-known UC Merced Land Use dataset we obtain a gain of almost 3\% with respect to the best reference,
and almost 5\% on the more recent Brazilian Coffee Scenes dataset.
Note that CNNs had been already applied to remote sensing scene classification, in \cite{Penatti2015}.
In that work, however, only the output of the penultimate layer of pre-trained networks is used, as a ``shallow'' image descriptor.

In the rest of the paper,
after a thorough analysis of related work in the field (Section II),
we provide the necessary background on ConvNets (Section III),
describe how we applied them to the scene classification problem (Section IV),
comment experimental results, also in comparison with reference techniques (Section V)
and finally draw conclusions (Section VI).

\section{Related work}

In the last few years, there has been intense research on remote sensing scene classification,
focusing both on the use of suitable image descriptors and of a proper classification task.
Local descriptors, in fact, like local binary patterns (LBP) \cite{Ojala2002},
scale-invariant feature transform (SIFT) \cite{Lowe2004}, or histograms of oriented gradients (HOG) \cite{Dalal2005},
with their invariance to geometric and photometric transformations, have proven effective in a variety of computer vision applications,
especially object recognition.
They can be extracted both in sparse (keypoint-based) and dense way.
In any case, given the high dimensionality of the feature space,
they need a subsequent coding phase in order to obtain an expressive but compact representation of the image.

The bag of visual words (BOVW) is a common and successful tool to reach this goal.
In its basic version, k-means clustering is used to create off-line a dictionary of visual words.
This is then used on-line to quantize the extracted features and associate with each one the label of the closest cluster centroid.
Eventually, the histogram of such labels is fed to a classifier, typically a support vector machine (SVM).
The SIFT-BOVW approach has been successfully applied in \cite{Yang2010} to land use image classification for remote sensing applications.

The basic version of BOVW, however, neglects information on the spatial distribution of visual words.
Hence, there have been several efforts in the literature to make up for this deficiency.
One popular approach is the spatial pyramid match kernel (SPMK) proposed in \cite{Lazebnik2006} for object and scene categorization.
It consists in partitioning the image at different levels of resolution
and computing weighted histograms of the number of matches of local features at each level.
Another alternative, considered in \cite{Jiang2012}, is to perform a randomized spatial partition (RSP),
aiming at a better characterization of the spatial layout of the images.
These partition patterns are then weighted according to their discriminative abilities, and boosted into a robust classifier.

Note that SPMK considers only the absolute spatial arrangement of visual words.
In order to capture both their absolute and relative spatial arrangements
the spatial co-occurrence kernel (SCK) \cite{Yang2010} and its pyramidal version (SPCK) \cite{Yang2011} were proposed.
Improved versions can be obtained by simply combining SCK and SPCK with the BOVW model (SCK+BOVW and SPCK+, respectively)
or SPCK with SPMK (SPCK++) \cite{Yang2010, Yang2011}.
The same goal of capturing both absolute and relative spatial relationships of local features is pursued in \cite{Chen2015},
where a pyramid-of-spatial-relatons (PSR) is proposed, which achieves higher robustness to rotations and translations.
A major difference with respect to previous approaches is that SIFT features are evaluated densely and not only on interest points.

Dense SIFT features are used also in \cite{Cheriyadat2014},
and encoded in terms of a learnt basis functions to generate a new sparse representation for the feature descriptors.
In \cite{Zhang2013}, instead, histograms of dense SIFT features are matched through a fast approximation of the Earth mover's distance.
This helps exploring the relations among visual codes,
which can be used as a key discriminative feature for image classification.
Another way to improve the histogram-based feature extraction is proposed in \cite{Kobayashi2014},
where histograms are regarded as Dirichlet-distributed probability mass functions,
and then transformed through a Fisher kernel to enhance their discriminative power.
Strong improvements also come from using more recent encoding methods as done in \cite{Negrel2014}, where
Fisher vectors (FV) \cite{Perronnin2010},
vectors of locally aggregated descriptors (VLAD) \cite{Jegou2011},
and vectors of locally aggregated tensors (VLAT) \cite{Picard2013}
are used in combination with HOG and color features.

Another effective way to improve the classification performance
is to augment the available dataset by adding rotated or flipped versions of the training images.
For this reason \cite{Xie2015} proposes Max-SIFT,
a flipping invariant descriptor which is obtained from the maximum of a SIFT descriptor and its flipped copy.

Most of the features proposed in the current literature are extracted only from the gray-scale image.
However, Yang and Newsam \cite{Yang2010} showed that
color histogram descriptors, evaluated on hue, lightness and saturation (Color-HLS), may provide a very good performance.
The interactions among RGB color bands is exploited also in mCENTRIST \cite{Xiao2014},
an extension of the CENTRIST algorithm \cite{Wu2011}, based in turn on LBP histograms and principal component analysis.
LBP features are used also in \cite{Ren2015},
where the maximal conditional mutual information (MCMI) scheme is proposed to select an optimal subset of these features.
In \cite{Chen2015a}, instead, feature selection is performed separately for each class,
with both one-versus-all and one-versus-one strategies.

Another trend is towards the combination of different features, pursued for example in \cite{Risojevic2013, Shao2013, Zhong2015}.
In \cite{Shao2013}, in particular,
a hierarchical scheme for multiple feature fusion (HMFF) is proposed.
In order to capture structural, shape, textural and color characteristics, four different features are considered,
and a two-step classification procedure is performed.

In \cite{Cheng2014} the use of high-level features is advocated for complex real-world scenes recognition.
A visual parts-based method is proposed, inspired by \cite{Felzenszwalb2010},
where several part detectors are used for objects or patterns at various orientations.
Improvements of this method are presented in \cite{Cheng2015} and in \cite{Cheng2015a}.
High-level features are used also in \cite{Yang2015} in the context of semisupervised feature learning.
\cite{Cheriyadat2014} resorts instead to fully unsupervised feature learning (UFL),
and the same does \cite{Hu2015}, together with spectral clustering (UFL-SC).

We conclude this review by mentioning the very recent work of \cite{Penatti2015},
the only one, to the best of our knowledge, considering ConvNets for land use classification.
As already said, however, CNNs are only used to produce shallow feature vectors for SVM classification,
and no training on remote sensing data is carried out.
Nonetheless, the performance is competitive with the previous state of the art.

\section{Convolutional Neural Networks}

\begin{figure}[!t]
\centerline{\includegraphics[width=10cm]{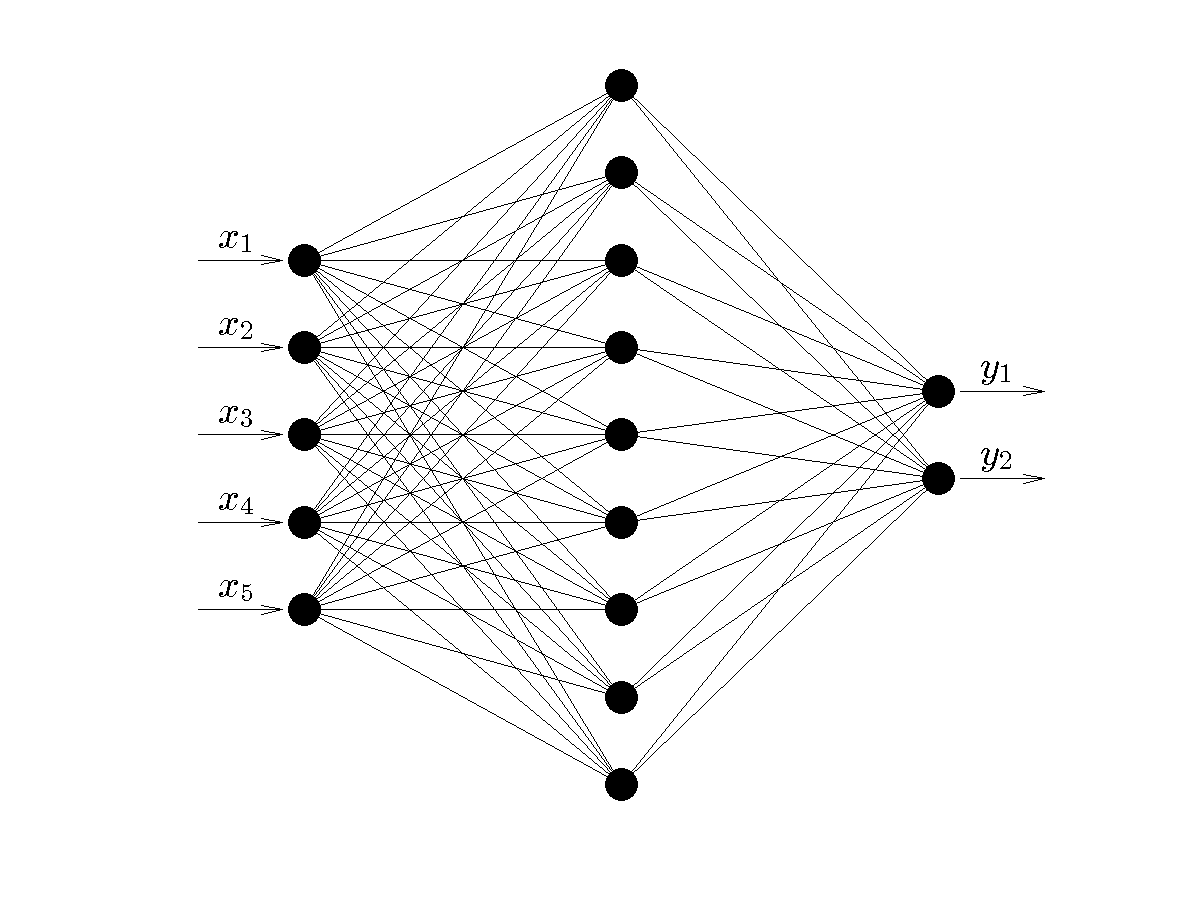}}
\caption{An example three-layer Perceptron.
This architecture may be used to make a binary decision on a 5-component input vector, using 9 features computed in the hidden layer.
The same basic architecture, with 784, 30 and 10 neurons in the input, hidden and output layers, respectively, can be used to classify handwritten digits with accuracy exceeding 97\%.}
\label{fig:example_MLP}
\end{figure}

The interest for convolutional neural networks (CNN) has been growing very fast, in the last few years,
because of their impressive results \cite{Krizhevsky2012}
in a series of challenging problems involving image classification and retrieval \cite{LeCun2015}.
CNN's evolve from the multilayer perceptron (MLP), proposed back in 1974 \cite{Werbos1974},
with a number of technical solution that help solve its bottlenecks.
Indeed, the MLP provides already excellent results in some classification problems.
For example,
a simple three-layer architecture similar to that of Fig.\ref{fig:example_MLP},
provides an accuracy beyond 97\% in the recognition of the handwritten digits of the MNIST dataset \cite{LeCun1998}.

\begin{figure}[!t]
\centerline{\includegraphics[width=8cm]{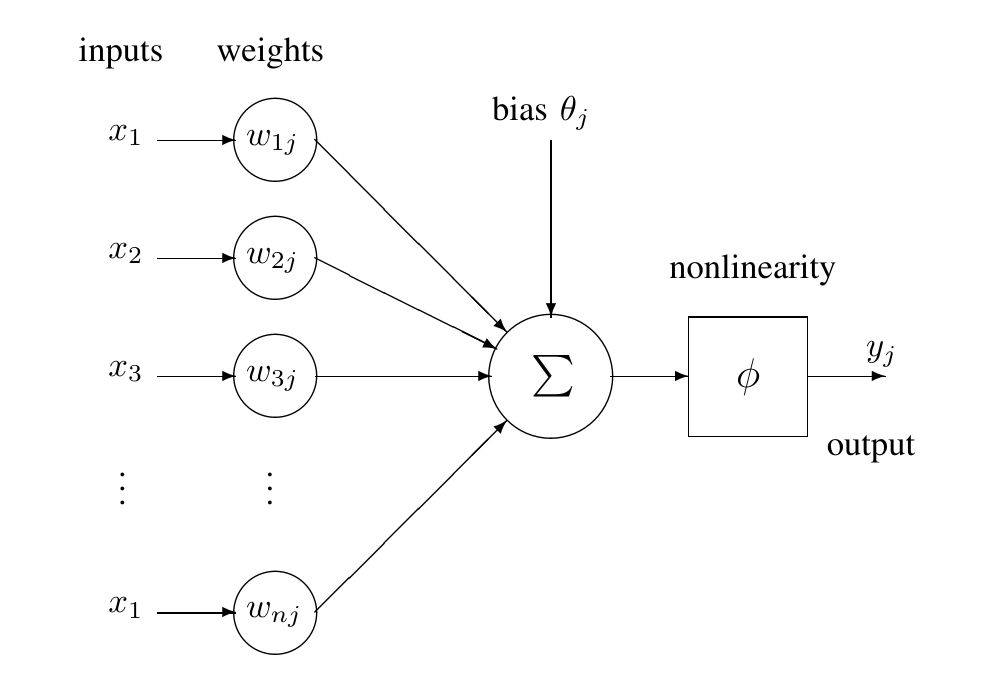}}
\caption{The neuron, computes a nonlinear function of the weighted sum of the inputs plus a bias term.}
\label{fig:neuron}
\end{figure}

The basic intuition behind these systems
is that a processing architecture based on a large number of layered and massively interconnected simple units,
may be more fit than sophisticated algorithms to tackle complex pattern recognition problems.
The basic processing unit, the neuron, is indeed very simple, as shown in Fig.\ref{fig:neuron}.
It computes the output activation by comparing the weighted sum of its input with a threshold and applying a suitable nonlinearity
\begin{equation}
    o_j = \phi(\sum_i w_{ij} x_i - \theta_j)
\end{equation}
Such configurable weights, $w_{ij}$ are the core of the net,
and they are learnt, typically through backpropagation \cite{Werbos1990}, based on an adequately large set of labeled training examples.

The main problem with MLP is the very large number of free parameters (the weights) to be set.
To solve challenging problems, such as image classification and retrieval, several layers of densely interconnected neurons must be considered.
Learning all these weights would require a huge training set, and exceedingly large computational power.
These problems are mostly solved by CNNs.

\begin{figure}[!t]
   \centerline{\includegraphics[width=8cm]{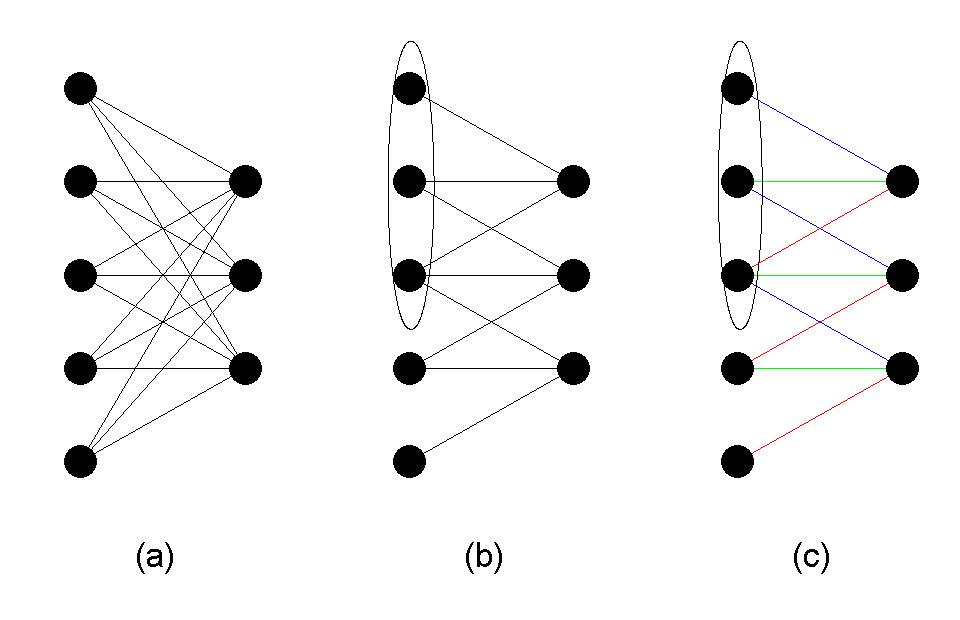}}
\caption{From MLP to CNN.
In MLP (a) all neurons of the second layer are fully connected with those of the first layer;
with CNNs, neurons have a limited receptive field, see the oval in (b);
moreover, all neurons of a layer share the same weights, see the color coding in (c).
In this toy example, the number of free parameter to learn drops from 15 to 3.}
\label{fig:MLP_vs_CNN}
\end{figure}

Convolutional neural networks were first proposed in 1980 by Fukushima \cite{Fukushima1980} (called NeoCognitron) and then refined by LeCun \cite{LeCun1989}.
Fig.\ref{fig:MLP_vs_CNN} provides some intuition into their major structural differences with respect to MLP.
While in MLP (a) all neurons of layer $(n+1)$ are connected to all neurons of layer $n$,
in CNN (b) neurons have limited ``receptive fields''.
Moreover, all neurons of a layer are identical to one another, except for their receptive fields, sharing the same weights, color coded in (c).
These constraints reduce sharply the number of free parameters to learn.
As the name suggests, the $(n+1)$-th layer computes (before the nonlinearity) a spatial convolution of the outputs of the $n$-th layer.
Therefore, it extracts some basic features of the image which are passed on to the next layer for further processing.

Besides being simpler than MLP, this architecture is much more similar to its biological template.
In fact,
according to the studies initiated by Hubel and Wiesel \cite{Hubel1959},
the visual cortex is organized in layers composed by similar cells, with different receptive fields over the lower layer.
In the lowest layer, elementary features are extracted, corresponding to bright spots, lines, corners, etc.,
which are then combined in higher layers so as to match more and more complex templates.

\begin{figure}[!t]
\centerline{\includegraphics[width=8cm]{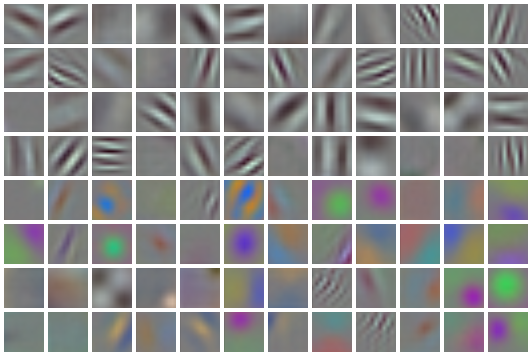}}
\caption{The first-layer filters learned by CaffeNet over general purpose optical images.
Filters appear to be sensitive to low-pass blobs and band-pass patterns of various frequencies and orientations, using both intensity and color information.}
\label{fig:firstlayer_filters}
\end{figure}

In actual CNNs,
each layer comprises various sublayers of neurons operating in parallel on the previous layer,
so as to extract a number of features at once, like a bank of filter does.
An example is given in Fig.\ref{fig:firstlayer_filters},
which shows the filters learned in the first layer of a CNN designed for image classification.
Each filter elaborates the three input color bands, producing by convolution a corresponding feature map.
In this case, therefore, 96 feature maps are produced as output of the second layer, which become the input of the next one.

Although CNNs have been introduced many years ago,
only in the last few years implementing and training large CNNs has become possible due to both technology progresses and research findings.
Major enabling factors have been the fast growth of affordable computing power, especially graphical processing units (GPUs),
and the diffusion of large datasets of labeled images for training.
On the other hand,
research has contributed many new solution for faster and more reliable learning,
like the rectified linear unit (ReLU) \cite{Jarrett2009, Nair2010}, a neuron with a simplified non-linearity which allows much faster training,
and techniques to reduce over-fitting, such as dropout \cite{Hinton2012}, or data augmentation.
The first CNN exploiting all these solutions, proposed in the seminal work of Krizhevsky {\it et al.} \cite{Krizhevsky2012},
improved image classification results by more than 10\% w.r.t. the previous state of the art.

Today's CNN architectures comprise typically several layers, of various types.
\subsubsection{Convolutional layers}
described before, are the most important ones.
They compute the convolution of the input image with the weights of the network.
Neurons in the first hidden layer view only a small image window, and learn low-level features.
Those in deeper layers view (indirectly) larger portions of the image, and are able to learn more expressive features by combining low-level ones. Each layer is characterized by a few hyper-parameters:
the number of filters to learn, their spatial support, the stride between different windows
and an optional zero-padding which controls the size of the layer output.
\subsubsection{Pooling layers}
reduce the size of the input layer through some local non-linear operations, for example $\max()$,
so as to reduce the number of parameters to learn and provide some translation invariance.
The most relevant hyper-parameters are the support of the pooling window and the stride between different windows.
\subsubsection{Normalization layers}
inspired by inhibition schemes present in the real neurons of the brain, aim at improving generalization.
They are typically used with sigmoid neurons (not ReLU).
\subsubsection{Fully-connected layers} are typically used as the last few layers of the network.
By removing constraints, they can better summarize the information conveyed by lower-level layers in view of the final decision.
Despite full connectivity, their complexity is still affordable thanks to the previous size-reducing layers.

\section{Using CNNs for remote sensing scene classification}

In this work, we use Convolutional Neural Networks to carry out remote sensing scene classification.
Several architectures have been already proposed and tested in the literature \cite{Chatfield2014}, especially for computer vision tasks,
and most of them have been implemented and made available online.
Here we focus on two very promising architectures, CaffeNet and GoogLeNet.

\begin{figure*}[!t]
\centerline{\includegraphics[width=16cm]{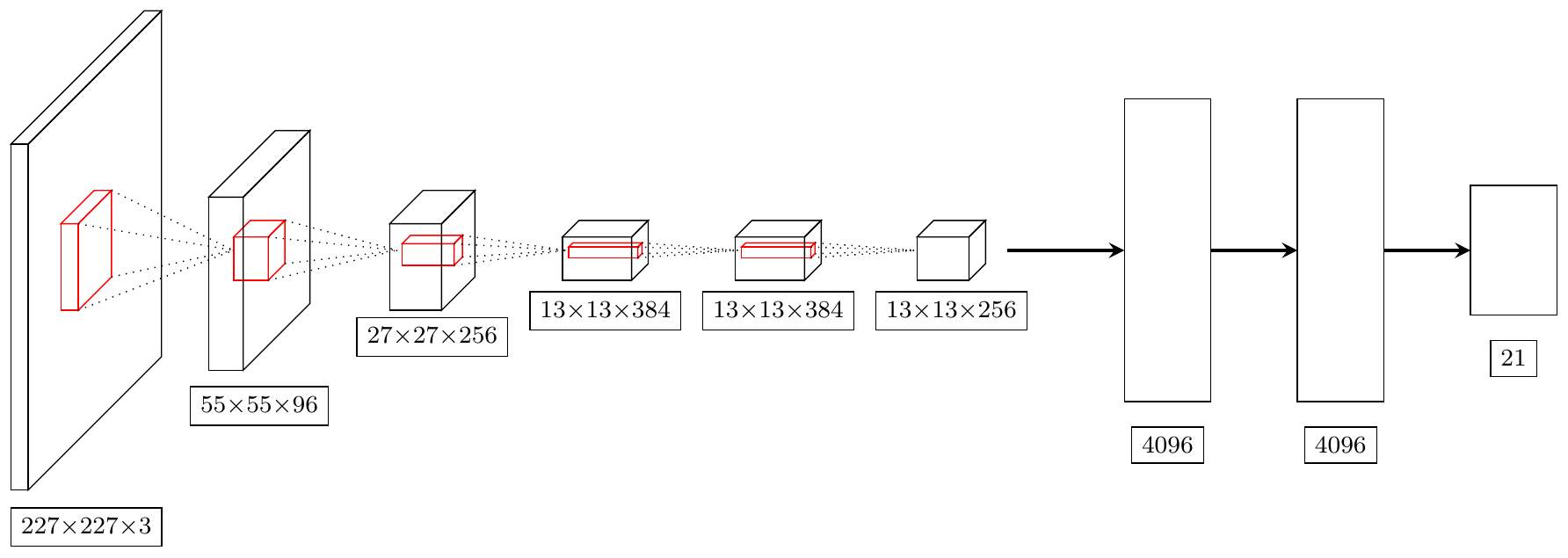}}
\caption{The CaffeNet architecture used in this work.
The boxes show the size of each feature layer and, for fully connected layers, the size of the output.
Most receptive fields are 3$\times$3, maxpool layers are not shown.}
\label{fig:caffenet}
\end{figure*}

Caffe is one of the most popular libraries for deep learning (convolutional neural networks in particular).
It is developed by the Berkeley Vision and Learning Center (BVLC) and community contributors.
Caffe is easily customizable through configuration files,
easily extendible with new layer types,
and provides a very fast ConvNet implementation (leveraging GPUs, if present).
It provides C++, Python and MATLAB APIs.
Fig.~\ref{fig:caffenet} shows the specific architecture used in this work,
the reference implementation from \cite{Jia2014}, which is in turn a modification of the network from \cite{Krizhevsky2012}
(the reader is referred to the original papers for more details).
It comprises 5 convolutional layers, each followed by a pooling layer, and 3 fully-connected layers.

\begin{figure}[!t]
\centerline{\includegraphics[width=8cm]{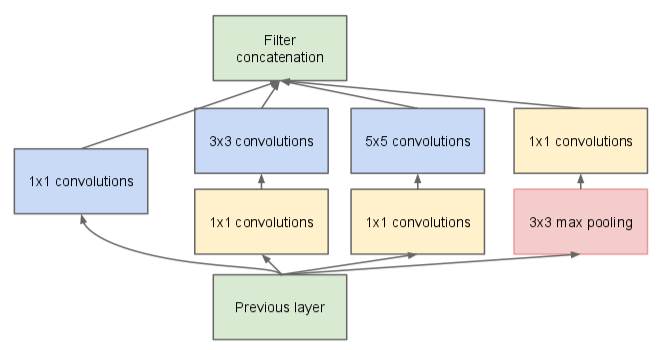}}
\caption{The inception module; convolutions of different sizes allow the network to process features at different spatial scales.
They are then aggregated and fed to the next layer.
1x1 convolutions are used for dimension reduction before the more expensive 3x3 and 5x5 convolutions.}
\label{fig:inception}
\end{figure}

GoogLeNet, presented in \cite{Szegedy2015}, is the CNN architecture that won the ILSVRC14 competition.
Its main peculiarity is the use of ``inception modules'',
based on the ``network in network`` idea of \cite{Lin2014}, 
inspired also by theoretical results from \cite{Arora2014}. 
In very concise terms,
inception modules reduce the complexity of the expensive 3d filters of conventional architectures by means of a prior depth reduction phase.
Thanks to the reduced complexity, multiple filters can be use in parallel at different resolutions, as shown in Fig.~\ref{fig:inception}.
To improve the effectiveness of the gradient backpropagation, given the depth of the network,
GoogLeNet employs also auxiliary classifiers connected to intermediate layers.
The Inception architecture has two main advantages:
{\it i)}  by employing filters of different sizes at each layer, it retains more accurate spatial information; moreover
{\it ii)} it significantly reduces the number of free parameters of the network,
          making it less prone to overfitting and allowing it to be deeper.
The latter is a critical feature for performance, according to the findings from \cite{Simonyan2014}. 
We refer the reader to the original paper by Szegedy {\it et al.} \cite{Szegedy2015}
for a detailed graphical illustration of the 22-layer GoogLeNet architecture with all relevant parameters.

These architectures have been developed to process natural images for computer vision applications.
In that context,
huge datasets of labeled images have been created and made available online,
in particular Imagenet\footnote{{\tt http://www.image-net.org/challenges/LSVRC/}} \cite{Russakovsky2015},
associated with a score of object classification challenges.
When trying to use these CNNs on the available remote-sensing datasets, we run into the major problem of limited training data.
In fact,
even the largest datasets available in this field are, to date, far too small for correctly training a large CNN.
The typical effect is overtraining,
that is, the network works perfectly on the training data but does not generalize well to test data, providing eventually poor results.

This is a problem common to many other tasks where training data are hard to obtain.
A possible solution, already explored in the literature, e.g. \cite{Razavian2014, Donahue2013, Yosinski2014},
is to use a pre-trained CNN and repurpose it to the task of interest.
Features learned in the lower layers of a CNN, in fact, like edges or color blobs,
may be general enough to be useful for other classification tasks as well.
Clearly, the success of this approach depends on several factors,
the most important being the ``distance'' between the original task on which the CNN was originally trained and the target task.
In particular, using CNNs trained on the Imagenet dataset makes full sense for UC-Merced data,
since optical remote-sensing images have strong low-level similarities with general-purpose optical images.
The same would not apply to SAR images, for example, due to their peculiar pixel-level statistics.
In the experimental section we will also consider a borderline case, with a dataset of remote sensing images including an infrared band.
As a further non-negligible advantage,
the fine-tuning approach is usually faster than training the CNN from scratch, which may take days or weeks of computation time.

\begin{figure*}
\centering
\begin{tabular}{ccccccc}
\includegraphics[width=.26\columnwidth]{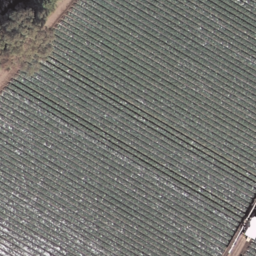}\sep
\includegraphics[width=.26\columnwidth]{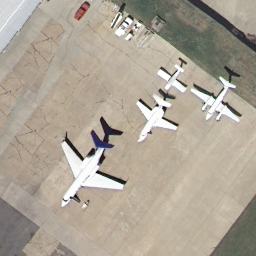}\sep
\includegraphics[width=.26\columnwidth]{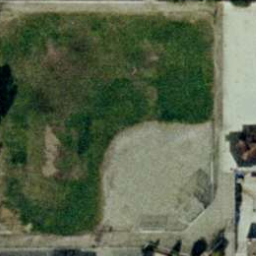}\sep
\includegraphics[width=.26\columnwidth]{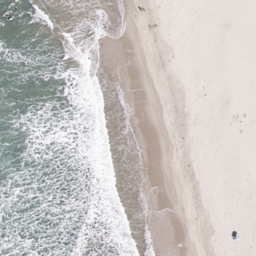}\sep
\includegraphics[width=.26\columnwidth]{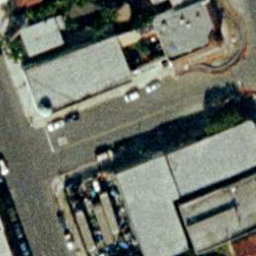}\sep
\includegraphics[width=.26\columnwidth]{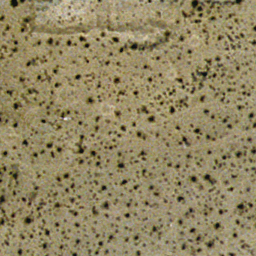}\sep
\includegraphics[width=.26\columnwidth]{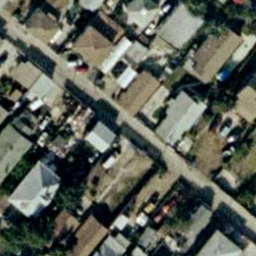}\\
(a)&(b)&(c)&(d)&(e)&(f)&(g)\\
\includegraphics[width=.26\columnwidth]{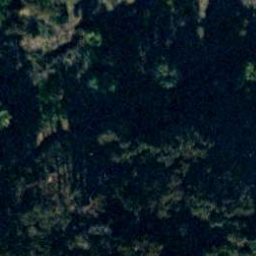}\sep
\includegraphics[width=.26\columnwidth]{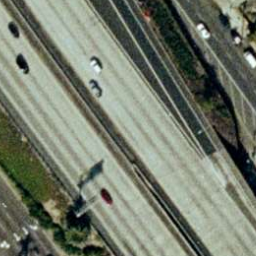}\sep
\includegraphics[width=.26\columnwidth]{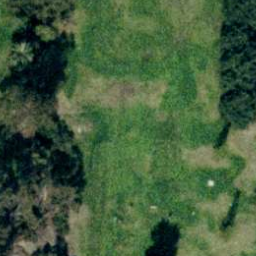}\sep
\includegraphics[width=.26\columnwidth]{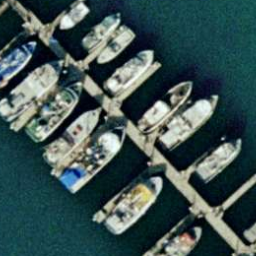}\sep
\includegraphics[width=.26\columnwidth]{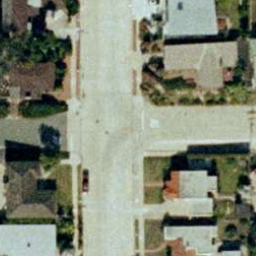}\sep
\includegraphics[width=.26\columnwidth]{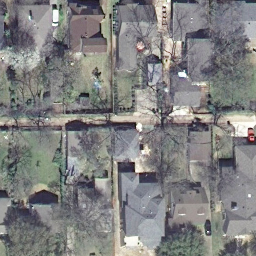}\sep
\includegraphics[width=.26\columnwidth]{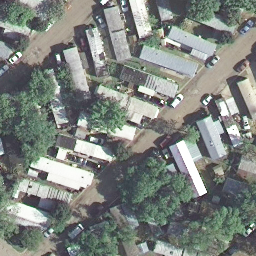}\\
(h)&(i)&(j)&(k)&(l)&(m)&(n)\\
\includegraphics[width=.26\columnwidth]{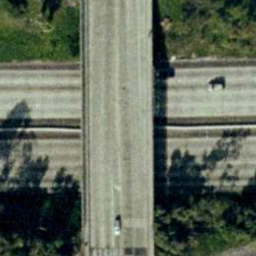}\sep
\includegraphics[width=.26\columnwidth]{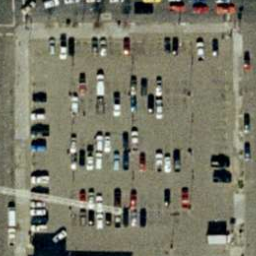}\sep
\includegraphics[width=.26\columnwidth]{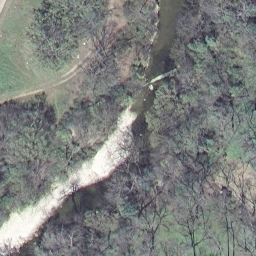}\sep
\includegraphics[width=.26\columnwidth]{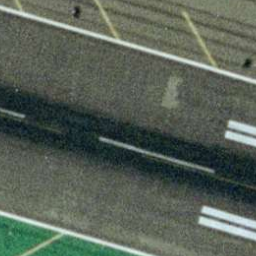}\sep
\includegraphics[width=.26\columnwidth]{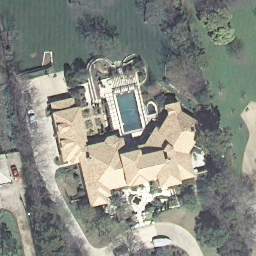}\sep
\includegraphics[width=.26\columnwidth]{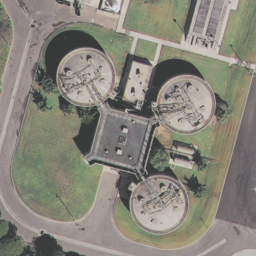}\sep
\includegraphics[width=.26\columnwidth]{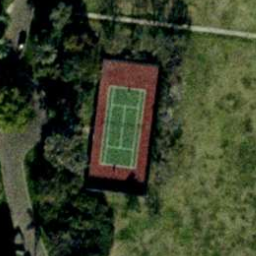}\\
(o)&(p)&(q)&(r)&(s)&(t)&(u)\\
\end{tabular}
\caption{Class representatives of the UC-Merced dataset.
(a) agricultural; (b) airplane; (c) baseball diamond; (d) beach; (e) buildings; (f) chaparral; (g) dense residential;
(h) forest; (i) freeway; (j) golf course; (k) harbor; (l) intersection; (m) medium residential; (n) mobile home park;
(o) overpass; (p) parking lot; (q) river; (r) runway; (s) sparse residential; (t) storage tanks; (u) tennis court;
}
\label{fig:UCMerced_images}
\end{figure*}

There are two major ways to adapt a pre-trained CNN
\begin{enumerate}
\item   submitting the training images to the CNN, and regarding the output of the penultimate layer as feature vectors, used to train an off-line classifier;
\item   use the training images to ``fine-tune'' the CNN for the task of interest.
\end{enumerate}
The first solution is straightforward and does not require any further effort, except for the design of a classifier.
It has been already considered for the classification of aerial images \cite{Penatti2015}, with good results.
Nonetheless, the second solution is certainly more promising,
as it allows a deeper adaptation to the data of interest, exploiting the full potential of CNNs.
With fine-tuning, one needs to decide which layers of the original network must be freezed, and which ones are instead allowed to keep learning, and at which rate.
These choices impact on both accuracy and design time and, as before,
depend very much on the similarity between the original and target problem, and on the amount of training data available.
Typically, the first few layers are freezed, because low-level features can better fit different problems.

In summary, in this work, we consider and compare three options:
\noindent
\begin{itemize}
\item   {\em training from scratch:} the whole convolutional net is trained on the available target data;
\item   {\em fine tuning:} a pre-trained net is used, adapting only a certain number of high-level layers;
\item   {\em feature vector:} the output of the penultimate layer of a pre-trained net is used as a feature vector for classification.
\end{itemize}
Note, however, that, differently from \cite{Penatti2015},
we implement the last option by replacing the last layer of the network  with a different fully-connected layer, with 21 outputs instead of 1000,
followed by a Softmax classifier instead of an SVM.
By so doing, we use only CNNs for all our experiments.

\section{Experimental results}

We carried out a number of experiments to assess the performance of the proposed approach,
also in comparison with state-of-the-art results.
We use two remote-sensing datasets.
The well-known UC Merced Land Use dataset \cite{Yang2010} (UC-Merced for short),
includes aerial optical images, with low-level characteristics similar to those of the Imagenet.
In recent years, many researchers have used this dataset, allowing for an extensive comparison of results with the literature.
The Brazilian Coffee Scenes dataset \cite{Penatti2015}, instead, includes satellite images with an infra-red band, hence less similar to general purpose images.
Since it has been published very recently, more limited results are available, including however results with CNNs.
In the next two subsections we discuss results separately for the two datasets

Experiments have been all carried out on a notebook equipped with an NVIDIA GeForce GT 750M 2048 MB GPU.
For each training modality and dataset, the number of training iterations has been established by preliminary experiments.
In fine-tuning modality, all learning rates are equal except for the first layer's, set to one tenth of the others.
In feature vector modality, only the last fully connected layer is trained.
In all cases we used a moderate data augmentation, through mirroring and random cropping, to increase the effective training set size.

\subsection{UC-Merced}

This dataset\footnote{{\tt http://vision.ucmerced.edu/datasets/landuse.html}}, released in 2010 \cite{Yang2010},
is extracted from large optical images (RGB color space) of the US Geological Survey, taken over various regions of the United States.
2100 256$\times$256-pixel images are selected and manually labeled as belonging to 21 land use classes, 100 for each class.
Fig.\ref{fig:UCMerced_images} shows one example image for each class.
More examples and more information are available in the original paper \cite{Yang2010}.
Due to their nature and relatively high resolution (30 cm) these images share many low-level features with general-purpose optical images
making them good candidates for fine-tuning a pre-trained CNN.

\renewcommand{\ru}{\rule{0mm}{4mm}}
\begin{table}[t]
\centering
{\small
\begin{tabular}{|c|c|c|c|}\hline
                 CNN           & Design         & Iterations &   Accuracy \ru \\ \hline \hline
\multirow{3}{*}{\ru CaffeNet}  & from scratch   &    100,000 &     85.71  \ru \\ \cline{2-4}
                               & fine-tuning    &     20,000 &     95.48  \ru \\ \cline{2-4}
                               & feature vector &      5,000 &     94.28  \ru \\ \hline \hline
\multirow{3}{*}{\ru GoogLeNet} & from scratch   &    100,000 &     92.86  \ru \\ \cline{2-4}
                               & fine-tuning    &     20,000 & \bf{97.10} \ru \\ \cline{2-4}
                               & feature vector &      5,000 &     94.38  \ru \\ \hline
\end{tabular}
}
\vspace{2mm}
\caption{Classification accuracy (\%) of proposed CNN-based solutions on the UC-Merced dataset. Best result in bold.}
\label{tab:UCMerced_proposed}
\end{table}

\begin{table*}[ht]
\centering
{\small
\begin{tabular}{|l|r|c|c|l|}
\hline
\ru Method        &        Ref.           &  Year  & Accuracy & Approach                                              \\ \hline \hline
\ru BOVW          &                       &        & 76.81    & Keypoint SIFT and Bag-of-Visual-Words                 \\ \hline
\ru SPMK          &   \cite{Lazebnik2006} &  2006  & 75.29    & Keypoint SIFT and Spatial Pyramid Match Kernel        \\ \hline
\ru Color-HLS     &       \cite{Yang2010} &  2010  & 81.19    & Hue, Lightness, Saturation and Bag-of-Visual-Words    \\ \hline
\ru SCK           &       \cite{Yang2010} &  2010  & 72.52    & Keypoint SIFT and Spatial Co-occurrence Kernel        \\ \hline
\ru BOVW+SCK      &       \cite{Yang2010} &  2010  & 77.71    & Keypoint SIFT, Spatial Co-occurrence Kernel, and Bag-of-Visual-Words   \\ \hline
\ru SPCK          &       \cite{Yang2011} &  2011  & 73.14    & Keypoint SIFT and Spatial Pyramid Co-occurrence Kernel \\ \hline
\ru SPCK+         &       \cite{Yang2011} &  2011  & 76.05    & Keypoint SIFT, Spatial Pyramid Co-occurrence Kernel, and Bag-of-Visual-Words \\ \hline
\ru SPCK++        &       \cite{Yang2011} &  2011  & 77.38    & Keypoint SIFT, Spatial Pyramid Match Kernel, and Spatial Pyramid Co-occurrence Kernel                       \\ \hline
\ru BRSP          &      \cite{Jiang2012} &  2012  & 77.80    & Randomized Spatial Partition via Boosting \\ \hline
\ru HMFF          &       \cite{Shao2013} &  2013  & 92.38    & Multi-Feature Fusion and Hierarchical classifier \\ \hline
\ru Dirichlet     &  \cite{Kobayashi2014} &  2014  & 92.80    & SIFT and Dirichlet-based Histogram Feature Transform \\ \hline
\ru UFL           & \cite{Cheriyadat2014} &  2014  & 81.67    & dense SIFT and Unsupervised Feature Learning \\ \hline
\ru mCENTRIST     &       \cite{Xiao2014} &  2014  & 89.90    & LBP on RGB + PCA            \\ \hline
\ru FV            &     \cite{Negrel2014} &  2014  & 93.80    & HOG + RGB and Fisher Vectors \\ \hline
\ru VLAD          &     \cite{Negrel2014} &  2014  & 92.50    & HOG + RGB and Vectors of Locally Aggregated Descriptors  \\ \hline
\ru VLAT          &     \cite{Negrel2014} &  2014  & 94.30    & HOG + RGB and Vectors of Locally Aggregated Tensors     \\ \hline
\ru COPD          &      \cite{Cheng2014} &  2014  & 91.33    & Collection Of Part Detectors                          \\ \hline
\ru Partlets      &      \cite{Cheng2015} &  2015  & 91.33    & Partlets \\ \hline
\ru Sparselets    &     \cite{Cheng2015a} &  2015  & 91.46    & Sparselets  \\ \hline
\ru MCMI-based    &        \cite{Ren2015} &  2015  & 88.20    & LBP and feature selection based on incremental Maximal-Conditional-Mutual-Information\\ \hline
\ru PSR           &       \cite{Chen2015} &  2015  & 89.10    & dense SIFT, Pyramid-of-Spatial-Relatons, and Bag-of-Visual-Words \\ \hline
\ru UFL-SC        &         \cite{Hu2015} &  2015  & 90.26    & Unsupervised Feature Learning with Spectral Clustering and Bag-of-Visual-Words \\ \hline
\ru CNN           &    \cite{Penatti2015} &  2015  & 93.42    & pre-trained ConvNet (CaffeNet) with SVM classifier \\ \hline
\ru proposed      &                       &  2015 &{\bf 97.10}& pre-trained ConvNet (GoogLeNet) with fine-tuning on target data \\ \hline
\end{tabular}
}
\vspace{2mm}
\caption{Classification accuracy (\%) of reference and proposed methods on the UC-Merced dataset. Best result in bold.}
\label{tab:UCMerced_comparison}
\end{table*}

In Table \ref{tab:UCMerced_proposed} we report synthetic results for the two CNN architectures and the three design approaches considered.
Results are always computed through five-fold validation, by averaging over the five folds.
The first observation is that the fine-tuning approach, as expected, provides the best results with both CaffeNet and GoogLeNet,
reaching an overall accuracy of 95.48\% and 97.10\%, respectively.
This is about 10\% and 5\% better, respectively, than the design from scratch,
confirming the limited value of this latter option when training data are limited.
As for the feature vector approach,
pretty good results are obtained with both architectures, but clearly inferior to those of the fine-tuning approach, with a gap of 1-3\%.

In terms of computation time, the training from scratch is also much more demanding.
Through preliminary experiments, we decided the number of training iterations to use for each modality,
100,000 for the training from scratch, much less for the fine tuning (20,000), and still less for the feature vector case (5,000).
These are not marginal differences, for such computation-intensive experiments.
Notice, however, that the fine tuning approach provides a very good performance already at 5,000 iterations,
95.12\% and 96.48\% respectively, hence it is preferable anyway to the feature vector approach.

Turning to the comparison between the two CNN architectures,
GoogLeNet appears to provide consistently the best performance, as suggested by the literature,
while being slightly less demanding in terms of computation.
In the following, when talking of the proposed approach, we will hence refer to GoogLeNet with fine tuning over 20,000 iterations.

Several approaches have been proposed recently for remote sensing scene classification,
and most of them have been tested on the UC-Merced dataset, following the same experimental protocol, with 5-fold cross validation.
Therefore there is plenty of data available for a solid comparison with the state of the art.
In Table \ref{tab:UCMerced_comparison} we report the overall accuracies for all these comparable methods, as they appear in the original papers,
together with the accuracy of our best CNN solution.
The proposed method guarantees a large performance gain w.r.t. to all references, with a minimum gap of almost 3\%.
This applies also to \cite{Penatti2015} which uses feature vectors extracted from CaffeNet, with pre-training on Imagenet.
Notice that, in the very same conditions,
we obtain somewhat better results (see Table \ref{tab:UCMerced_proposed}) probably because of the different classifier.

\begin{figure*}[!t]
\centerline{\includegraphics[width=18cm]{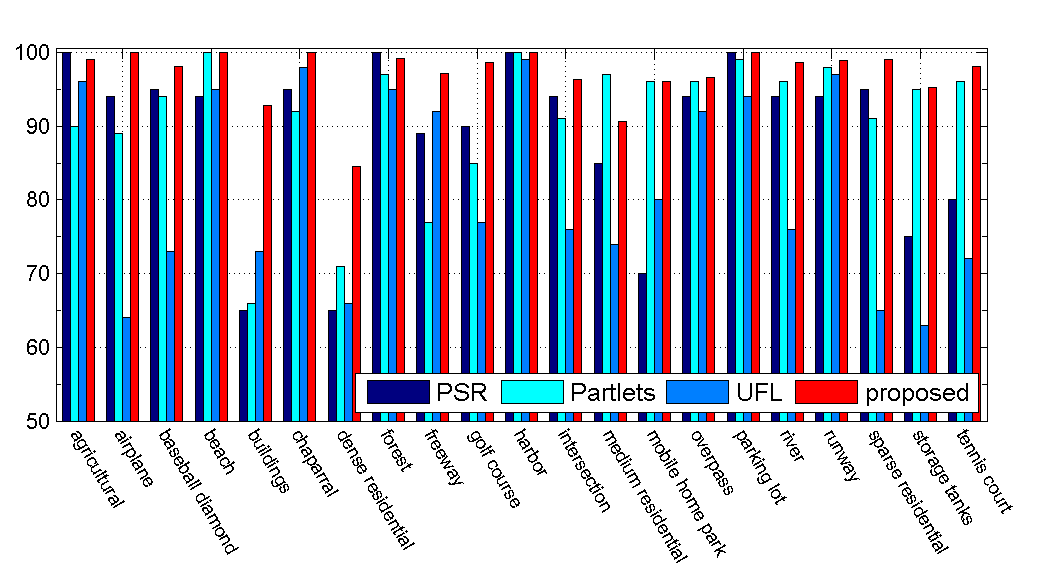}}
\caption{Per-class accuracies of the proposed method and some state-of-the-art references on the UC-Merced dataset.}
\label{fig:perclass_accuracy}
\end{figure*}

In Fig.\ref{fig:perclass_accuracy} we show the per-class accuracies provided by the proposed method and some selected references,
PSR \cite{Chen2015}, Partlets \cite{Cheng2015}, and UFL \cite{Cheriyadat2014},
using again the results reported in the original papers.
The proposed method provides the best performance almost uniformly over all classes, often with perfect or near-perfect accuracy.
The worst result (84.5\% accuracy) is observed for the dense residential class which, as already observed,
suffers the presence of other very close classes, like medium residential and mobile home park.
In any case, for the same class, reference methods perform more than 10\% worse.

\begin{figure}
\centering
\begin{tabular}{ccc}
\includegraphics[width=.26\columnwidth]{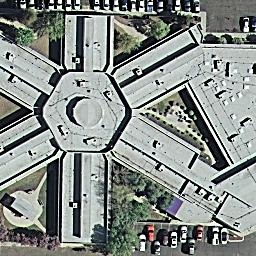}\sep
\includegraphics[width=.26\columnwidth]{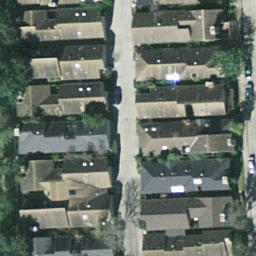}\sep
\includegraphics[width=.26\columnwidth]{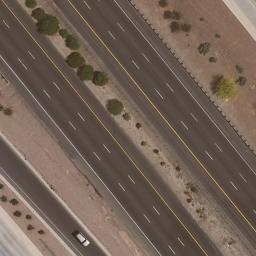}\\
(a)&(b)&(c)\\
\includegraphics[width=.26\columnwidth]{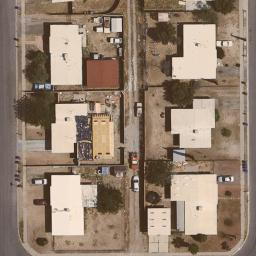}\sep
\includegraphics[width=.26\columnwidth]{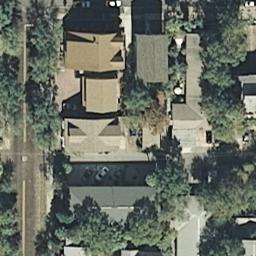}\sep
\includegraphics[width=.26\columnwidth]{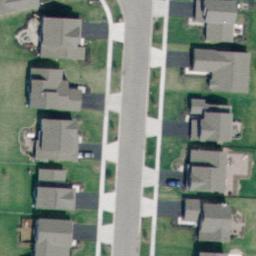}\\
(d)&(e)&(f)\\
\includegraphics[width=.26\columnwidth]{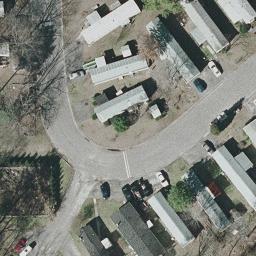}\sep
\includegraphics[width=.26\columnwidth]{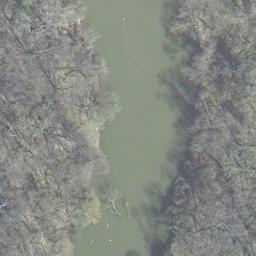}\sep\\
(g)&(h)&\\
\end{tabular}
\caption{Classification errors in fold \#1 of UC-Merced by the proposed method.
(a) buildings classified as ($\to$)storage tanks;
(b) dense residential $\to$ mobile home park;
(c) freeway $\to$ runway;
(d) medium residential $\to$ dense residential;
(e) medium residential $\to$ sparse residential;
(f) medium residential $\to$ dense residential;
(g) mobile home park $\to$ intersection;
(h) river $\to$ golf course;}
\label{fig:UCMerced_errors}
\end{figure}

Finally, in Fig.\ref{fig:UCMerced_errors}, we show all the images of fold \#1 (only one fold, to save space) that have been wrongly classified with the proposed method.
Only 8 images out of a total of 420 have been misclassified, in this fold,
all belonging to classes that have some very ``close'' neighbors, as noted commenting the bar graph of Fig.\ref{fig:perclass_accuracy}.
A correct classification of some of these images may be difficult also for a human photointerpreter.
It is worth underlining, however,
that a suitable fusion of the outputs of CaffeNet and GoogLeNet would remove virtually all these errors, as also observed in \cite{Penatti2015}.

\subsection{Brazilian Coffee Scenes}

\begin{figure}
\centering
\begin{tabular}{cccc}
\includegraphics[width=.20\columnwidth]{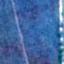}\sep
\includegraphics[width=.20\columnwidth]{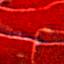}\sep
\includegraphics[width=.20\columnwidth]{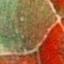}\sep
\includegraphics[width=.20\columnwidth]{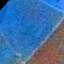}\\
(a)&(b)&(c)&(d)\\
\includegraphics[width=.20\columnwidth]{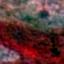}\sep
\includegraphics[width=.20\columnwidth]{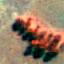}\sep
\includegraphics[width=.20\columnwidth]{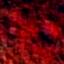}\sep
\includegraphics[width=.20\columnwidth]{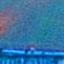}\\
(e)&(f)&(g)&(h)\\
\end{tabular}
\caption{Class representatives of the Brazilian Coffee Scenes dataset. (a)-(d) coffee tiles; (e)-(h) non-coffee tiles.}
\label{fig:brazil_images}
\end{figure}

This dataset\footnote{{\tt www.patreo.dcc.ufmg.br/downloads/brazilian-coffee-dataset/}}, released in 2015 \cite{Penatti2015},
includes scenes taken by the SPOT sensor in the green, red, and near-infrared bands, over four counties in the State of Minas Gerais, Brazil.
The scenes are partitioned in over 50,000 64$\times$64-pixel tiles, labeled as coffee (1,438) non-coffee (36,577) or mixed (12,989).
To provide a balanced dataset,
1,438 tiles of both coffee and non-coffee classes are kept and organized in 5 almost equally sized folds,
while mixed tiles are all discarded.
Fig.\ref{fig:brazil_images} shows four tiles each of the coffee and non-coffee classes, obviously in false colors.
Overall, this dataset is very different from UC-Merced.
The images are not optical (green-red-infrared instead of red-green-blue),
hence intrinsically different from the Imagenet samples used for pre-training.
Moreover, there is a much larger number of samples for each class, which could make the design from scratch a more interesting option for CNN.
In this case, however, there is no reference result for a comparative assessment of performance,
except for those reported in \cite{Penatti2015}.

\renewcommand{\ru}{\rule{0mm}{4mm}}
\begin{table}[t]
\centering
{\small
\begin{tabular}{|c|c|c|c|}\hline
                    CNN        & Design         & Iterations &  Accuracy \ru \\ \hline \hline
\multirow{3}{*}{\ru CaffeNet}  & from scratch   &     20,000 &    90.17  \ru \\ \cline{2-4}
                               & fine-tuning    &     10,000 &    90.94  \ru \\ \cline{2-4}
                               & feature vector &      5,000 &    85.02  \ru \\ \hline \hline
\multirow{3}{*}{\ru GoogLeNet} & from scratch   &     20,000 &\bf{91.83} \ru \\ \cline{2-4}
                               & fine-tuning    &     10,000 &    90.75  \ru \\ \cline{2-4}
                               & feature vector &      5,000 &    84.02  \ru \\ \hline
\end{tabular}
}
\vspace{2mm}
\caption{Classification accuracy (\%) of proposed CNN-based solutions on the Brazilian Coffee Scenes dataset. Best result in bold.}
\label{tab:CoffeePlantation_proposed}
\end{table}

Table \ref{tab:CoffeePlantation_proposed} shows the results obtained with the proposed techniques.
The most notable difference w.r.t. the UC-Merced case is the relatively poor performance of the feature vector approach.
This can be attributed to the more marked differences w.r.t. Imagenet dataset used to pre-train the network.
In fact, training from scratch provides a much better performance, in this case,
probably due also to the larger number of samples available for each class.
The optimal number of training iterations is much smaller than before, 20,000, and results almost equally good are obtained also at 10,000 iterations.
Similar considerations apply to the fine-tuning case, where the best results are obtained at 10,000 iterations.
The overall best, almost 92\%, is provided by GoogLeNet with training from scratch.
In general, results are significantly worse than with UC-Merced, despite the 2-class vs. 21-class problem.
Indeed, as the Authors of \cite{Penatti2015} note,
this is a rather challenging dataset, due to a large intra-class variability
{\em ``caused by different crop management techniques, different plant ages and/or spectral distortions and shadows''}.
In any case, this result is almost 5\% better than the top result reported in \cite{Penatti2015}, see Tab.\ref{tab:brazil},
obtained with BIC (Border-Interior Pixel Classification) a simple color descriptor,
and 7\% better than CNNs with the feature vector approach, for the mentioned reasons.

\begin{table}[!t]
\centering
{\small
\begin{tabular}{|l|c|l|}
\hline
\ru Method   & Accuracy   & Approach                        \\ \hline \hline
\ru BIC      &     87.0   & color descriptors               \\ \hline
\ru BOVW     &     80.5   & BOVW with dense SIFT features   \\ \hline
\ru CNN-1    &     84.8   & pre-trained CaffeNet + SVM      \\ \hline
\ru CNN-2    &     81.2   & pre-trained OverFeat + SVM      \\ \hline \hline
\ru proposed & \bf{91.8}  & GoogLeNet trained from scratch  \\ \hline
\end{tabular}
}
\vspace{2mm}
\caption{Classification accuracy (\%) of reference and proposed methods on the Brazilian Coffee Scene dataset.
All reference data from \cite{Penatti2015}. Best result in bold.}
\label{tab:brazil}
\end{table}

\section{Conclusions}

We have addressed the remote sensing scene classification task by resorting to convolutional neural networks.
Two promising architectures have been considered with three design modalities.
Experiments on two datasets with quite different properties have provided insightful information.

As expected, training a deep CNN from scratch is not always advisable with the limited-size datasets currently available in this field.
A valid alternative consists in using pre-trained CNN and adapting it to the target task.
However, a shallow adaptation, with the CNN used only as a feature vector generator for subsequent classification,
gives up much of the potential of this approach.
On the contrary, a careful fine-tuning, involving several layers of the architecture, provides very good results, in general.
Overall, the experimental evidence is definitely encouraging.
On the widespread UC-Merced dataset, the proposed method outperforms the best references technique by almost 3\%.
Moreover, it provides the best performance, by a wide margin, also on the more recent Brazilian Coffee Scenes dataset.
This latter dataset allowed us to study also the behavior of this approach when pre-training and target data differ significantly.

The near-perfect performance on aerial images
makes clear that the next big challenge is related to the classification of data acquired with other imaging modalities.
The Synthetic Aperture Radar, therefore, is certainly a field of great interest for future research,
drawing already the attention of several research groups \cite{Popescu2012, Bahmanyar2015}.
A recent work in this direction is \cite{Wang2015}, addressing automatic target recognition in SAR images.

\balance
\bibliographystyle{IEEEbib}
\bibliography{refs}

\end{document}